# A Feedback Integrated Web-Based Multi-Criteria Group Decision Support Model for Contractor Selection using Fuzzy Analytic Hierarchy Process

Abimbola H. Afolayan [1], Bolanle A. Ojokoh [2], and Adebayo O. Adetunmbi[3]

[1] Department of Information Systems, Federal University of Technology, Akure, Nigeria
**ahafolayan@futa.edu.ng**
[2] Department of Information Systems, Federal University of Technology, Akure, Nigeria
**baojokoh@futa.edu.ng**
[3] Department of Computer Science, Federal University of Technology, Akure, Nigeria
**aoadetunmbi@futa.edu.ng**

**Abstract.** The construction sector constitutes one of the most important sectors in the economy of any country. Many construction projects experience time and cost overruns due to the wrong choice of contractors. In this paper, the feedback integrated multi-criteria group decision support model for contractor selection was proposed. The proposed model consists of two modules; technical evaluation module and financial evaluation module. The technical evaluation module is employed to screen out the contractors to a smaller set of acceptable contractors and the functionality of the module is based on the Fuzzy Analytic Hierarchy Process (FAHP). The outputs of the technical evaluation module are fed into the financial evaluation module, taken into account the bid price criterion, the differences between the project owner's cost estimate and the bid prices are calculated. The contractor having the lowest value of such difference is awarded the contract. This research work introduced a consistency module. In the case of high inconsistency, feedback and recommendation models are introduced. The system implementation was done using HTML, CSS, JavaScript, and MongoDB. At the end of the selection procedure, contractor 5 ranked highest and is therefore recommended as the most responsive contractor for the award of the contract.

## 1 Introduction and Related Works

Many important decisions in society are made by groups of individuals such as committees, governing bodies, juries, business partners, teams, and families [1]. In most situations when a group of people is making a decision, they do not have only one objective; instead, they need to take into consideration several different points of view. Towards this, multi-criteria methods may be used to guide the analysis by specifying the criteria involved in the decision to suggest a priority of choices among the alternatives [2].

Timing becomes an important issue when more than one decision-maker is involved in a multi-criteria decision making (MCDM) problem. If the number of elements used to define the problem is large, collecting data from decision-makers, organizing, analyzing, synthesizing, and finally reaching a conclusion becomes a tremendous effort. A visual and interactive decision-making tool that is web-based, becomes an inevitable opportunity to assist in solving this problem [3]. Web-based applications are increasingly being used for multi-criteria group decision support environments [4]. Web-based applications offer many advantages such as the possibility of carrying out distributed decision-making processes where decision-makers cannot meet physically; timely delivery, secure information, and tools with a user-friendly interface.

One widely used MCDM method is the Analytic Hierarchy Process (AHP). AHP was proposed by [5]. The major characteristic of the AHP method is the use of pair-wise comparisons, which are used both to compare the alternatives with respect to the various criteria and estimate criteria weights [6]. [3] highlighted some advantages of the AHP method which are the relative ease with which it handles multiple criteria. AHP is quite easy for most decision-makers to understand and it can effectively handle both qualitative and quantitative data, among others. AHP, despite its popularity and simplicity in concept, it is not adequate to take into account the uncertainty associated with the mapping of one's perception to a number [7].

Fuzzy Theory proposed by [8] has existed now for several decades. Fuzzy logic itself has proven to be an effective MCDM method. Like other artificial intelligence methods, it has some advantages within uncertain, imprecise and vague contexts than AHP; it is similar to human judgments [7]. Combining Fuzzy methods with AHP is one approach for solving the complicated problems of AHP. This issue has attracted many researchers to apply Fuzzy AHP in different fields such as risk and disaster management [9]; work safety evaluation [10]; green initiatives in the fashion supply chain [11]; geographical information system application [12]; risk evaluation of information technology projects [13]; water management plans assessment [14]; and eco-environmental vulnerability assessment [15], among others.

Over the years in Nigeria, the selection of a contractor has been done manually from the process of inviting contractors to bid for projects to the selection of successful bidders. The traditional systems of procurement through manual modes suffer from various problems such as inordinate delays (approximately 4 to 6 months) in tender/order processing, heavy paperwork, multi-level scrutiny that consumes a lot of time, physical threats to bidders, the human interface at every stage, inadequate transparency and discretionary treatment in the entire tender process, among others [16]. These could be linked to poor methods and procedures of the selection of contractors. The correct decision-making method is therefore required for selecting the appropriate contractor for a construction project. Using a multi-criteria approach for evaluating contractors may help to solve these problems [17].

Fuzzy Analytic Hierarchy Process (Fuzzy AHP) has been applied in contractor and supplier selection problems by several researchers: [18] developed a contractor selection in MCDM context using fuzzy AHP, [19] proposed multi-criteria supplier selection using Fuzzy-AHP Approach: A case study of manufacturing company and [20] developed a Fuzzy AHP model for the selection of consultant contractor in bidding phase in Vietnam. [21] apply the Fuzzy Analytic Hierarchy Process (FAHP) approach for supplier selection in the textile industry in Pakistan. However, each has one limitation or the other. For instance, in [18], the sub-criteria for selecting contractors were not included; In [19], consistency ratio was not calculated; In [20], though consistency ratio was calculated, there was no feedback and recommendation mechanism developed in the model in [21], The research only considered the main criteria; the sub-criteria for supplier selection in the textile industry was not considered, which if used may have differently ranked the suppliers. This research is motivated by the need to address these limitations.

The overall motivation for this research is, therefore, the need to develop a feedback integrated web-based multi-criteria group decision support model for contractor selection that will rank the main criteria, sub-criteria, and alternatives using Fuzzy AHP. The proposed model can check the consistency of decision-makers during the pairwise comparison of Fuzzy AHP. In the case of high inconsistency, Feedback and recommendation models are introduced; this would generate advice to the decision-makers if they are not consistent enough in their judgments during the pairwise comparison of criteria and alternatives. The rest of the paper is structured as follows: Section 2 describes the proposed methodology; in Section 3 the proposed methodology is applied to contractor selection and results are provided; finally, Section 4 includes conclusions of the present work and directions of future work.

## 2 The Proposed Feedback Integrated Multi-Criteria Group Decision Support Model for Contractor Selection

This research identified and assessed the existing contractors' decision criteria used for selecting a contractor in Federal Universities in Nigeria. The construction of No 1 Three-in-One Lecture Theatre for Centre for Entrepreneurship at The Federal University of Technology, Akure was used as a case study. In this case study, the advertisement required each bidder to submit bidding documents that would be used by the decision-makers to assess the contractors' technical capability. Only fifteen contractors submitted the bidding documents. The bidding documents consist of fifteen requirements, out of which the first ten requirements are mandatory. None submission of any of the mandatory requirements implies automatic disqualification of the bidder. Based on this condition, six contractors were disqualified from further participation in the technical evaluation exercise, thereby remaining nine contractors that submitted all the mandatory requirements. The contractors were evaluated with respect to four (4) criteria and twenty-six (26) sub-criteria as shown in Fig. 4.

The proposed model for contractor selection consists of two modules; technical evaluation module and financial evaluation module. The functionality of the technical evaluation module is based on the Fuzzy Analytic Hierarchy Process (FAHP). At the first stage, contractors are evaluated and scored with respect to the technical evaluation criteria, thereafter, the contractors are ranked according to their weights. Subsequently, the outputs of the technical evaluation module are fed into the financial evaluation module. Here bid price criterion is taken into account. The differences between the project owner's cost estimate and the bid prices are calculated. The contractor having the lowest value of such difference is awarded the contract.

In this research work, a consistency module is introduced. This module checks the inconsistency in decision-makers' judgments' during pairwise comparisons when using the Fuzzy AHP. In the case of high inconsistency, the decision-makers are advised to reconsider their pairwise comparisons using the feedback mechanism. The architecture of the web-based multi-criteria group decision support system for contractor selection is shown in Fig.1.

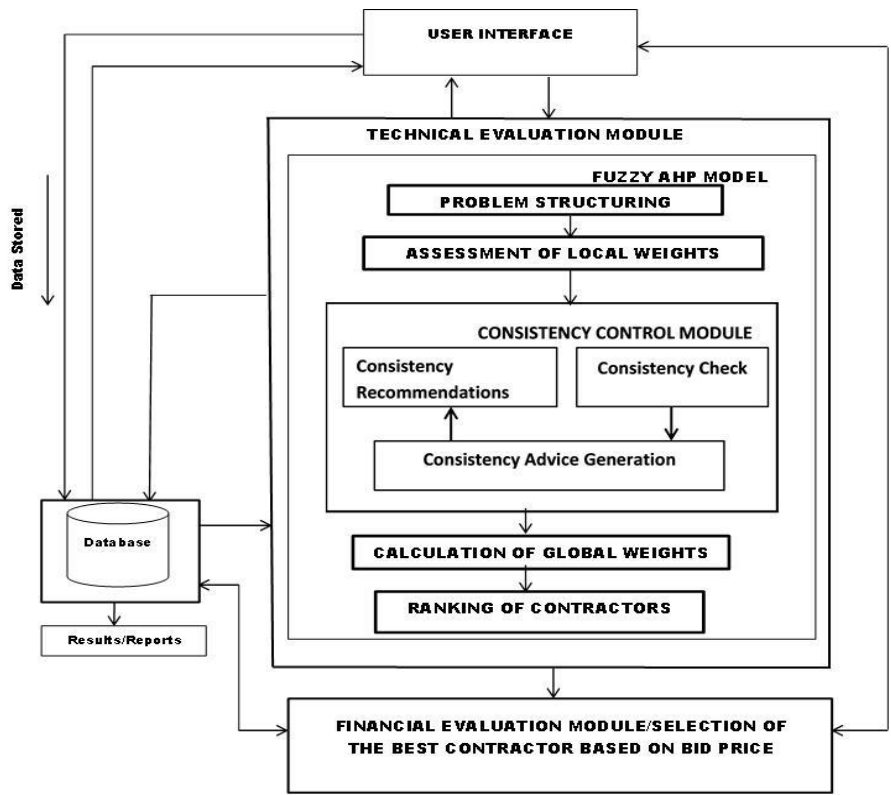

**Fig. 1.** Architecture of the Proposed Model

## 2.1 Technical Evaluation Module

The functionality of the technical evaluation module is based on the Fuzzy Analytic Hierarchy Process. The Fuzzy AHP divides the multi-criteria group decision-making problem into four main sub-modules: Problem structuring; Assessment of local weights; Consistency Control Module and Calculation of global weights.

### 2.1.1 Problem Structuring

The Fuzzy AHP problem structuring is described in the following steps:

**Step 1: Determine criteria, sub-criteria, and alternatives for the decision problem**

The following notations were used to describe the various entities in the proposed model for contractor selection:

Let

$$E = \{E_z\}, \forall\, z \in \{1, \ldots, 4\} \tag{1}$$

where $E$ is the list of decision-makers in the group and $e_z$ denotes individual group member

$$B = \{B_v\}, \forall\, v \in \{1, \ldots, 9\} \tag{2}$$

where $B$ is the list of contractors that bided for the contract and $B_v$ denotes individual contractor

$$C = \{C_n\}, \quad (n = 1, \ldots, 4) \tag{3}$$

where $n$ denote individual criteria

Let $C_{1j}$, $(j = 1, \ldots, 4)$ denotes sub-criteria in $C_1$; $C_{2k}$, $(k = 1, \ldots, 4)$ denotes sub-criteria in $C_2$; $C_{3m}$, $(m = 1, \ldots, 3)$ denotes sub-criteria in $C_3$; and $C_{4p}$, $(p = 1, \ldots, 4)$ denotes sub-criteria in $C_4$ respectively.

**Step 2: Establish a hierarchical structure**

The Fuzzy AHP decision problem is structured hierarchically at different levels, each level consisting of a finite number of decision elements. The hierarchical structure for selecting the best contractor consists of three levels. Level A, the target level, demonstrates the final objective of the whole hierarchical structure, which is selecting the best contractor. Level B contains the measurement criteria and sub-criteria. Level

C contains the contractors which are going to be measured and prioritized based on their performance. The hierarchical structure is shown in Fig. 2.

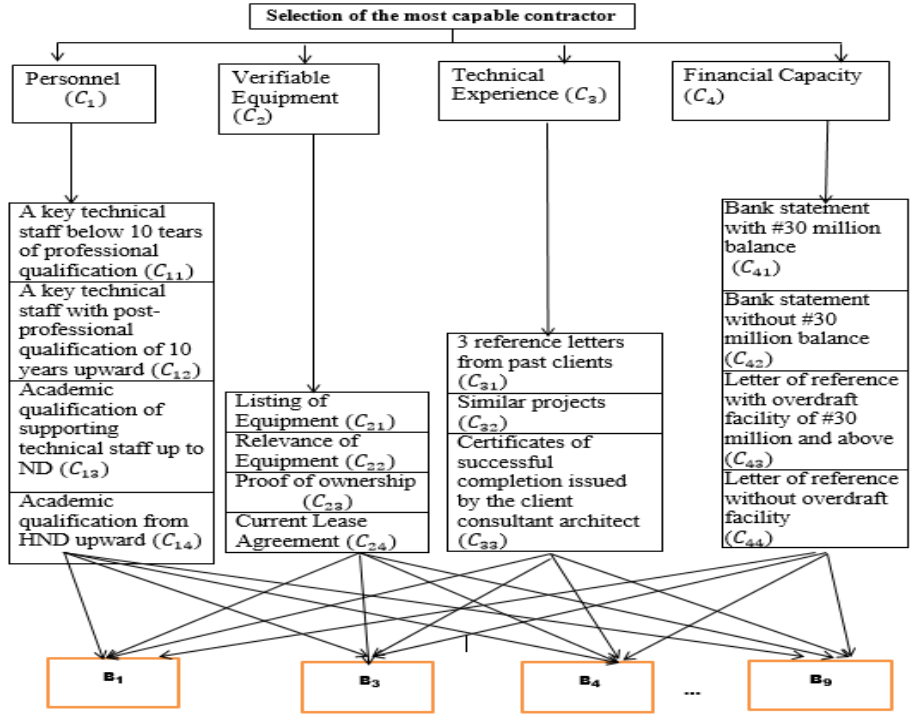

**Fig. 2.** The hierarchical structure of contractor selection

### Step 3: Collect experts' judgments based on fuzzy scale and establish fuzzy pair-wise comparison matrices of criteria, sub-criteria, and alternatives

In this step, each group member $e_z$ is required to fill up a pairwise comparison matrix of the relative importance of these criteria, sub-criteria, and alternatives using experts' judgments in the form of fuzzy numbers. The weights of the nine-level fundamental scales of judgments were expressed via triangular fuzzy numbers due to its simplicity and effectiveness [22]. Suppose that all pairwise comparison judgments are represented as fuzzy triangular numbers.

$$\tilde{a}_{ij} = (l_{ij}, m_{ij}, u_{ij}) \text{ such that } l_{ij} \leq m_{ij} \leq u_{ij} \qquad (4)$$

The membership function of the triangular fuzzy number $\tilde{a}_{ij}$ can be found in Eq. (5)

$$\mu_A(\tilde{a}_{ij}) = \begin{cases} \dfrac{\tilde{a}_{ij} - l_{ij}}{m_{ij} - l_{ij}} & l_{ij} \leq \tilde{a}_{ij} \leq m_{ij} \\ \dfrac{u_{ij} - \tilde{a}_{ij}}{u_{ij} - m_{ij}} & m_{ij} \leq \tilde{a}_{ij} \leq u_{ij} \\ 0 & \tilde{a}_{ij} < l_{ij} \text{ or } \tilde{a}_{ij} > u_{ij} \end{cases} \quad (5)$$

where $l_{ij}, m_{ij}, u_{ij}$ shows the minimum possible, most likely and the maximum possible value of a fuzzy number for $e^{th}$ decision maker's preference for $i^{th}$ criterion over $j^{th}$ criterion, via fuzzy triangular numbers, respectively.

The linguistic scale is given in Table 1 [23].

**Table 1.** Decision Maker compares the criteria via linguistic terms

| Linguistic Variables | Saaty's Scale | Fuzzy AHP Scale | |
|---|---|---|---|
| | | TFN | Reciprocal TFN |
| Equally Important | 1 | (1, 1, 1) | (1, 1, 1) |
| Equally to Moderately Important | 2 | (1, 2, 3) | (1/3, 1/2, 1) |
| Moderately Important | 3 | (2, 3, 4) | (1/4, 1/3, 1/2) |
| Moderately to Strongly Important | 4 | (3, 4, 5) | (1/5, 1/4, 1/3) |
| Strongly Important | 5 | (4, 5, 6) | (1/6, 1/5, 1/4) |
| Strongly to Very Strongly Important | 6 | (5, 6, 7) | (1/7, 1/6, 1/5) |
| Very Strongly Important | 7 | (6, 7, 8) | (1/8, 1/7, 1/6) |
| Very Strongly to Extremely Important | 8 | (7, 8, 9) | (1/9, 1/8, 1/7) |
| Extremely important | 9 | (8, 9, 9) | (1/9, 1/9, 1/8) |

The relative importance of the decision elements (weights of criteria, sub-criteria, and weights of alternatives) is assessed indirectly from comparison judgments during the second step of the decision process. The decision-maker is required to provide his/her preferences by comparing all criteria, sub-criteria, and alternatives with respect to upper-level decision elements.

Taking, for instance, the estimation of the importance of criteria, let $\tilde{A}$ represent the fuzzy judgment matrix of $e^{th}$ decision maker's preference for $i^{th}$ criterion over $j^{th}$ criterion, via fuzzy triangular numbers.

$$\widetilde{A} = [\widetilde{a}_{ij}] = \begin{array}{c} \\ C_1 \\ C_2 \\ \vdots \\ C_4 \end{array} \overset{\begin{array}{cccc} C_1 & C_2 & \cdots & C_4 \end{array}}{\begin{bmatrix} 1 & (l_{12}, m_{12}, u_{12}) & \cdots & (l_{14}, m_{14}, u_{14}) \\ \left(\frac{1}{u_{21}}, \frac{1}{m_{21}}, \frac{1}{l_{21}}\right) & 1 & \cdots & (l_{24}, m_{24}, u_{24}) \\ \vdots & \vdots & \vdots & \vdots \\ \left(\frac{1}{u_{i4}}, \frac{1}{m_{14}}, \frac{1}{l_{14}}\right) & \left(\frac{1}{u_{24}}, \frac{1}{m_{24}}, \frac{1}{l_{24}}\right) & \cdots & 1 \end{bmatrix}} \quad (6)$$

where $\widetilde{a}_{ij} = 1: \forall i = j;\ \widetilde{a}_{ji} = \frac{1}{\widetilde{a}_{ij}}: \forall i \neq j$, $\widetilde{a}_{ij}$ is the fuzzy evaluation between criterion $i$ and criterion $j$ of decision-maker $e_z$.

The same step would be taken for the estimation of sub-criteria in respect to each criterion, for instance, the relative importance of sub-criteria $(C_{11}, C_{12}, C_{13}, C_{14})$ with respect to criterion $(C_1)$ is shown in Eq. (7) and the relative importance of contractors that bid for the contract with respect to sub-criterion $(C_{11})$ is: shown in Eq. (8):

$$[\widetilde{a}_{ij}] = \begin{array}{c} \\ C_{11} \\ C_{12} \\ \vdots \\ C_{14} \end{array} \overset{\begin{array}{cccc} C_{11} & C_{12} & \cdots & C_{14} \end{array}}{\begin{bmatrix} 1 & (l_{12}, m_{12}, u_{12}) & \cdots & (l_{14}, m_{14}, u_{14}) \\ \left(\frac{1}{u_{21}}, \frac{1}{m_{21}}, \frac{1}{l_{21}}\right) & 1 & \cdots & (l_{24}, m_{24}, u_{24}) \\ \vdots & \vdots & \vdots & \vdots \\ \left(\frac{1}{u_{14}}, \frac{1}{m_{14}}, \frac{1}{l_{14}}\right) & \left(\frac{1}{u_{24}}, \frac{1}{m_{24}}, \frac{1}{l_{24}}\right) & \cdots & 1 \end{bmatrix}} \quad (7)$$

$$[\widetilde{a}_{ij}] = \begin{array}{c} \\ B_1 \\ B_2 \\ \vdots \\ B_y \end{array} \overset{\begin{array}{cccc} B_1 & B_2 & \cdots & B_y \end{array}}{\begin{bmatrix} 1 & (l_{12}, m_{12}, u_{12}) & \cdots & (l_{1y}, m_{1y}, u_{1y}) \\ \left(\frac{1}{u_{21}}, \frac{1}{m_{21}}, \frac{1}{l_{21}}\right) & 1 & \cdots & (l_{2y}, m_{2y}, u_{2y}) \\ \vdots & \vdots & \vdots & \vdots \\ \left(\frac{1}{u_{iy}}, \frac{1}{m_{iy}}, \frac{1}{l_{iy}}\right) & \left(\frac{1}{u_{2y}}, \frac{1}{m_{2y}}, \frac{1}{l_{2y}}\right) & \cdots & 1 \end{bmatrix}} \quad (8)$$

**Step 4: Aggregate the Experts' Judgments**

To aggregate the experts' judgments of criteria, sub-criteria and alternatives, [24]'s method is applied and this is shown in Eq. (9).

$$\tilde{r}_i = \left(\prod_{i=1}^{4}(\widetilde{a}_{ij})\right)^{1/4} \quad (9)$$

where $\tilde{r}_i$ is the geometric mean of fuzzy comparison values

### 2.1.2 Assessment of Local Weights

The following steps are used to calculate the local weights of each criterion, sub-criterion, and alternatives.

**Step 1: Determine Fuzzy Weights of Criteria, Sub-criteria, and Alternatives**

To calculate the fuzzy weights of each criterion, sub-criterion, and alternatives, [25]'s model is adopted. From Eq. (9), three sub-steps are incorporated:

**Step a**: Find the vector summation of each $\tilde{r}_i$.

$$\sum_i^4 \tilde{r}_i = \tilde{r}_{i1} \oplus \tilde{r}_{i2} \oplus \ldots \oplus \tilde{r}_{i4} \qquad (10)$$

**Step b**: Find the (-1) power of the summation vector. Replace the fuzzy triangular number, to make it an increasing order.

$$(\sum_i^4 \tilde{r}_i)^{-1} = (\tilde{r}_{i1} \oplus \tilde{r}_{i2} \oplus \ldots \oplus \tilde{r}_{i4})^{-1} \qquad (11)$$

**Step c**: To find the relative fuzzy weight of criterion $i$, multiply each $\tilde{r}_i$ with this increasing order of the reverse vector.

$$\tilde{w}_i = \tilde{r}_i \otimes (\tilde{r}_{i1} \oplus \tilde{r}_{i2} \oplus \ldots \oplus \tilde{r}_{i4})^{-1} \qquad (12)$$
$$= l\tilde{w}_i, m\tilde{w}_i, u\tilde{w}_i$$

where $\tilde{w}_i$ are the relative fuzzy weights of criteria, sub-criteria and contractors and $l\tilde{w}_i$ is the minimum possible value and $m\tilde{w}_i$ is the most likely value and, $u\tilde{w}_i$ is the maximum possible value of a fuzzy number.

**Step 2: Defuzzify and Normalize the Relative Fuzzy Weights of Criteria, Sub-criteria, and Alternatives**

Since $\tilde{w}_i$ are still fuzzy triangular numbers, they need to be de-fuzzified. To defuzzify the fuzzy triangular numbers, the centre of area method proposed by [26] is applied via Eq. (13).

$$M_i = \frac{l\tilde{w}_i \oplus m\tilde{w}_i \oplus u\tilde{w}_i}{3} \qquad (13)$$

where $M_i$ is the relative non-fuzzy weight of each criterion, sub-criterion, and contractors.

To normalize $M_i$, matrix $M_i$ is transformed into matrix $V = [v_i]$, the elements of matrix $V$ are calculated according to Eq. (14).

$$v_i = \frac{M_i}{\sum_{i=1}^{4} M_i} \tag{14}$$

where $v_i$ is the normalized non-fuzzy weights.

**Step 3: Determine the local weights of criteria, sub-criteria, and alternatives**

The local weights $[w_i]$ are calculated via Eq. (15).

$$w_i = \frac{v_i}{4} \quad \forall\, i \in \{1, \dots, 4\} \tag{15}$$

### 2.1.3 Consistency Control Module

The consistency control module consists of two steps: consistency ratio computation and the feedback process to generate advice and recommendations.

**Step 1: Computing the Consistency Ratio**

[27] model is adopted in calculating the Consistency Ratio (CR). CR is a comparison between the Consistency Index and Random Index as shown in Eq. (16).

$$CR = \frac{CI}{RI} \tag{16}$$

where $CI$ is the consistency Index and $RI$ is the Random Consistency Index

$$CI = \frac{\lambda_{max} - n}{n - 1} \tag{17}$$

where $n$ is the number of rows in the decision matrix and $\lambda_{max}$ is the largest eigenvalue

Judgments that have a CR lower than 0.1 are acceptable (Saaty, 1980) and higher than 0.1 are not acceptable.

To calculate the $\lambda_{max}$, [28] model is adopted as shown in Eq. (18)

$$\lambda_{max} = \sum_{i=1}^{4} \frac{(Aw_i)}{w_i} \tag{18}$$

where $A$ = matrix operating on the ratio of $w_i$ and $w_j$, $w_i$ are the local weights

$A$ is expressed as:

$$A = \frac{w_i}{w_j} \quad \forall \, i,j \in \{1,2,\ldots,4\} \tag{19}$$

Therefore:

$$Aw_i = \begin{bmatrix} \frac{w_1}{w_1} & \frac{w_1}{w_2} & \cdots & \frac{w_1}{w_4} \\ \frac{w_2}{w_1} & \frac{w_2}{w_2} & \cdots & \frac{w_2}{w_4} \\ \vdots & \vdots & & \vdots \\ \frac{w_4}{w_1} & \frac{w_4}{w_2} & \cdots & \frac{w_4}{w_4} \end{bmatrix} \begin{bmatrix} w_1 \\ w_2 \\ \vdots \\ w_4 \end{bmatrix} \tag{20}$$

From equation (19), RI depends on the size of the matrix $n$. The values of (RI) of sample size 10 as established by Saaty (2012) are given in Table 2.

**Table 2.** Random Consistency Index (RI)

| $n$ | 1 | 2 | 3 | 4 | 5 | 6 | 7 | 8 | 9 | 10 |
|---|---|---|---|---|---|---|---|---|---|---|
| **RI** | 0 | 0 | 0.58 | 0.9 | 1.12 | 1.24 | 1.32 | 1.41 | 1.45 | 1.49 |

**Step 2: Feedback Process to Generate Advice and Recommendations**

In the case of high inconsistency, the decision-makers are advised to reconsider their pairwise comparisons using the feedback mechanism, until the consistency measure is below the threshold ($\gamma$) indicated. [29] model is adapted in developing the feedback mechanism. The production of advice is carried out using two kinds of rules: identification rules and direction rules.

1. Identification Rules (IR): Here, the decision-maker $e_z$, whose $CR > \gamma$ or otherwise is identified. In this stage, there are two identification rules, they are Identification rule 1 (IR.1) and identification rule 2 (IR.2). (IR.1) identifies the set of pairwise comparison ($\tilde{a}_{ij}$) of decision-makers $e_z$, whose consistency ratio is greater than the consistency threshold value($\gamma$), and (IR.2) identifies the set of pairwise comparison ($\tilde{a}_{ij}$) of decision-makers $e_z$, whose consistency ratio falls between the range 0 and 0.1.

Let the consistency threshold ($\gamma$) be $0 \leq \gamma \leq 0.1$

$$IR = \begin{Bmatrix} IR.1 & if \ CR > \gamma \ \forall \ (\tilde{a}_{ij} \in e_z) \\ IR.2 & if \ CR \leq \gamma \ \forall \ (\tilde{a}_{ij} \in e_z) \end{Bmatrix} \tag{21}$$

where ($\tilde{a}_{ij} \in e_z$) denotes pairwise comparisons $\tilde{a}_{ij}$ that belongs to decision-maker $e_z$

2. Direction Rule (DR): Direction rule is applied to find out the direction of the change to be recommended in each case of the identification rule. The following direction rules are defined:

$$DR = \begin{Bmatrix} DR.\,1: & \text{If CR} > \gamma \text{ Reject and modify the fuzzy pairwise comparison} \\ DR.\,2: & \text{If CR} \leq \gamma \text{ Accept} \end{Bmatrix} \quad (22)$$

### 2.1.4 Calculation of Final Global Weights and Ranking of Contractors with respect to Sub-criteria

The global weights of the contractor are calculated by aggregating all local weights generated from the first, second and third levels of the hierarchical structure through the simple weighted sum.

Assume the local weights generated for each criterion by decision-maker $e_z$ are: $wC_1$ is the local weight assigned to the criterion $C_1$, $wC_2$ is the local weight assigned to the criterion $C_2$, $wC_3$ is the local weight assigned to the criterion $C_3$, $wC_4$ is the local weight assigned to the criterion $C_4$. Also, assume the local weights generated by decision-maker $e_z$ for sub-criteria $(C_{11}, C_{12}, C_{13}, C_{14})$ with respect to criterion $(C_1)$ are $(wC_{11}, wC_{12}, wC_{13}, wC_{14})$.

Therefore, the global weights for the sub-criteria $(C_{11}, C_{12}, C_{13}, C_{14})$ with respect to $C_1$ is calculated via Eq. (26).

$$WC_{1j} = w\,C_1 \times wC_{1j} \quad \forall\, j \in \{1, \ldots, 4\} \quad (23)$$

where $WC_{1j}$ are the global weights of sub-criteria; $wC_1$ is the local weight of criterion 1
$wC_{1j}$ are the local weights of sub-criteria

Considering the third level of the hierarchical structure. The local weights generated for each contractor with respect to each sub-criterion, taking sub-criterion $(C_{11})$ for instance are: $wC_{11}B_1$ is the local weight decision maker $e_z$ assigned to contractor $B_1$; $wC_{11}B_2$ is the local weight decision maker $e_z$ assigned to contractor $B_2$ and $wC_{11}B_3 = $ local weight decision maker $e_z$ assigned to contractor $B_3$ and so on.

The global weights for contractor $(B_1, B_2, B_3)$ with respect to sub-criterion $(C_{11})$ are:

$$WC_{11}B_v = WC_{11} \times wC_{11}B_v \quad \forall\, v \in [1, \ldots, 9] \quad (24)$$

where $WC_{11}B_v$ are the global weights assign to contractor ($B_v$) based on sub-criterion $C_{11}$, $WC_{11}$ are the global weights of sub-criterion $C_{11}$ and $wC_{11}B_v$ is the local weight decision maker $e_z$ assign to contractor $B_v$.

Generally, assume:

$WC_{1j}B_v$ are the global weights assigned to a contractor ($B_v$) based on sub-criterion in $C_{1j}$
$WC_{2k}B_v$ are the global weights assigned to a contractor ($B_v$) based on sub-criterion in $C_{2k}$
$WC_{3m}B_v$ are the global weights assigned to a contractor ($B_v$) based on sub-criterion in $C_{3m}$
$WC_{4p}B_v$ are the global weights assigned to a contractor ($B_v$) based on sub-criterion in $C_{4p}$

Therefore, the global weight of the contractor ($WB_v$) for all sub-criteria can be calculated via Eq. (25).

$$WB_v = \frac{\sum_{j=1}^{4} WC_{1j} \sum_{v=1}^{9} B_v + \sum_{k=1}^{4} WC_{2k} \sum_{v=1}^{9} B_v + \sum_{l=1}^{3} WC_{3m} \sum_{v=1}^{9} B_v + \sum_{m=1}^{4} WC_{4p} \sum_{v=1}^{9} B_v}{n} \quad (25)$$

where $B_v$ denotes individual contractor and $n$ is the total number of sub-criteria

Since there are more than one decision-makers, the global weight of each decision-maker in respect to each contractor ($WB_v e_z$) are averaged to get the final global weights of each alternative and is calculated via Eq. (26).

$$FWB_v = \frac{\sum_{z=1}^{4} WB_v e_z}{z} \quad (26)$$

where $e_z$ denotes individual decision-maker, z is the total number of decision-makers, and $FWB_v$ are the final global weights of contractors

The final global weights thus obtained are used for the final ranking of the contractors. Subsequently, after the ranking, contractors that are not technically satisfactory are further screen out, therefore in this research;

Let $\sigma$ be the threshold value for screening out contractors and;

$$FWB_v = \{FWB_1, FWB_2, \dots, FWB_v\} \ \forall \ v \in \{1, \dots, 9\} \quad (27)$$

Let $FWB_{v\,max}$ be the maximum final global weights of alternative;

$$\sigma = 50\% \times FWB_{v\,max} \quad (28)$$

where $\sigma = 0.5 FWB_{v\,max}$, $FWB_v \geq \sigma$ qualify for the financial evaluation module while $FWB_v < \sigma$ are screened out.

In this case study only seven contractors qualified for the financial evaluation, two contractors were screened out by the technical evaluation module.

## 2.2 Financial Evaluation Module

At the financial evaluation module, Topcu's (2004) model is adapted, where only the bid price criterion is taken into account. The differences between the project owner's cost estimate and the bid prices are calculated. The contractor having the lowest value of such difference is awarded the contract.

$$C_v = \{C_1, C_2, \ldots, C_u\} \ \forall \ (C_v \leq B_v), (C_v \in B_v) \tag{29}$$

where $C_v$ is the list of shortlisted contractors and $B_v$ is the list of contractors that bid for the contract

Let $d$ represents the Individual contractor's bid price

$$F_d = \{F_1, F_2, \ldots, F_q\} \ \forall \ d \in \{1, \ldots, 7\} \tag{30}$$

where $F_d$ is the list of the shortlisted contractors' bid price

Let $P_e$ be the consultant's estimated price, $\delta_v$ be the difference between the estimated price and bid price of shortlisted contractors

$$\delta_v = P_e - F_d \ \forall \ d \in \{1, \ldots, 7\} \tag{31}$$

Then,

$$\alpha = Min \ (\delta_v) \quad \forall \ (\alpha \in F_d) \tag{32}$$

where $\alpha$ denotes the best alternative bid price to be awarded the contract

## 3 Results and Discussions

To achieve the system robustness, flexibility, and resistance to potential change, the implementation of the proposed model is carried out in the Windows Operating System environment using HTML, CSS, and JavaScript at the presentation tier, JavaScript at the application tier and MongoDB at the data tier. Each decision-maker $e_z$ is required to register his/her account and login through the user interface of the web-based system.

The decision-makers' preferences are collected based on the Fuzzy AHP preference scale in Table 1 and the Fuzzy AHP pairwise comparison matrix of criteria is established according to Eq. 9, 10 and 11.

### 3.1 Local Weights Criteria for all Decision Makers

The local weights of each criterion for all decision-makers were aggregated and ranking of criteria was performed based on obtained values and the results are shown in Table 3.

**Table 3.** Sorting and Ranking of Local Weights of Criteria for Decision Makers

| Criteria | Aggregated Local weights of Criteria for all Decision Makers | Ranking |
|---|---|---|
| Technical Experience ($C_3$) | 0.432022 | 1 |
| Personnel ($C_1$) | 0.316947 | 2 |
| Verifiable Equipment($C_2$) | 0.177259 | 3 |
| Financial Capacity ($C_4$) | 0.073772 | 4 |

According to the result findings in Table 3, it can be deduced that Technical Experience ($C_3$) ranked first; thereafter, Personnel($C_1$) ranked second, followed by Verifiable Equipment and Financial capacity ranked last.

To check if the pairwise comparison matrix of criteria of decision-makers is consistent, taking, for instance, the results of the Consistency Ratio of decision maker1according to Eq. (16), Lambda Max according to Eq. (18), Consistency Index according to Eq. (17) are displayed in Table 4. The feedback recommendation displayed acceptable according to Eq. (22) since all considered fuzzy matrices are consistent and their consistency ratio is *CR* < 0.1.

**Table 4.** Consistency Status of Pairwise Comparison of Criteria by Decision Maker1

| | |
|---|---|
| Lambda Max | 4.004 |
| Random Index | 0.9 |
| Consistency Index | 0.0013333333333331865 |
| Consistency Ratio | 0.0014814814814813183 |
| **Since Consistency Ratio is < $0.1$, Status: Acceptable** ||

### 3.2 Global Weights of Sub-Criteria for all Decision Makers

The global weights of each sub-criterion for all decision-makers were aggregated and the criteria were ranked, the results are shown in Table 5.

**Table 5.** Sorting and Ranking of Global Weights of Sub-Criteria for Decision Makers

| Criteria | Sub-criterion | Aggregated Global weights of Sub-Criteria for all Decision Makers | Ranking |
|---|---|---|---|
| **Personnel** ($C_1$) | $C_{12}$ | 0.048173976 | 1 |
| | $C_{14}$ | 0.016060073 | 2 |
| | $C_{11}$ | 0.011425567 | 3 |
| | $C_{13}$ | 0.003577112 | 4 |
| **Verifiable Equipment** ($C_2$) | $C_{22}$ | 0.054085118 | 1 |
| | $C_{24}$ | 0.011922394 | 2 |
| | $C_{23}$ | 0.00763463 | 3 |
| | $C_{21}$ | 0.007264998 | 4 |
| **Technical Experience** ($C_3$) | $C_{32}$ | 0.091682049 | 1 |
| | $C_{31}$ | 0.034820556 | 2 |
| | $C_{33}$ | 0.01750465 | 3 |
| **Financial Capacity** ($C_4$) | $C_{41}$ | 0.011032273 | 1 |
| | $C_{42}$ | 0.002470265 | 2 |
| | $C_{43}$ | 0.002470265 | 2 |
| | $C_{44}$ | 0.002470265 | 2 |

According to the result findings in Table 5, it can be deduced that for Personnel criteria, a key technical staff with the post-professional qualification of 10 years upward ($C_{12}$) ranked first, while academic qualification from HND upward sub-criteria ($C_{14}$) ranked second, followed by a key technical staff below 10 years of post-professional qualification ($C_{11}$) and academic qualification of supporting technical staff up to ND ($C_{13}$) ranked last. Also for verifiable criterion, the relevance of equipment ($C_{22}$) ranked first, followed current lease of agreement ($C_{24}$) ranked second, while proof of ownership ($C_{23}$) ranked third and listing of equipment ($C_{21}$) ranked last. Likewise for the technical experience criteria, similar projects ($WC_{32}$) ranked first, followed by three reference letters from past clients ($WC_{31}$) and certificate of successful completion issued by the client/ consultant architect ($WC_{33}$) ranked last. Similarly for the financial capacity criterion, bank statement with #30 million ($WC_{41}$) ranked first and bank statement without #30 million balance ($WC_{42}$), letter of reference with overdraft facility of #30 million and above ($WC_{43}$) and letter of reference without overdraft facility ($WC_{44}$) ranked second.

### 3.3  Final Global Weights of Contractors for Each Decision Maker

The final global weights of contractors for each decision-maker as shown in Eq. (26) is depicted in Table 6.

**Table 6.** Final Global Weights of Contractors for all the Decision Makers

| Alternatives Names | DM. 1 Alternatives Global Weights | DM. 2 Alternatives Global Weights | DM. 3 Alternatives Global Weights | DM. 4 Alternatives Global Weights |
|---|---|---|---|---|
| Contractor 1 | 0.00096814 | 0.00077432 | 0.00085770 | 0.00270205 |
| Contractor 2 | 0.00031455 | 0.00068291 | 0.00053683 | 0.00064577 |
| Contractor 3 | 0.00099848 | 0.00075659 | 0.00088892 | 0.00280886 |
| Contractor 4 | 0.00124313 | 0.00072990 | 0.00097547 | 0.00254776 |
| Contractor 5 | 0.00093518 | 0.00080540 | 0.00084524 | 0.00047604 |
| Contractor 6 | 0.00134322 | 0.00109932 | 0.00114998 | 0.00083954 |
| Contractor 7 | 0.00029252 | 0.00030443 | 0.00044826 | 0.00028128 |
| Contractor 8 | 0.00134321 | 0.00110530 | 0.00123356 | 0.001038022 |
| Contractor 9 | 0.00100309 | 0.00104860 | 0.00085054 | 0.00048683 |

The ranking of the final global weights of alternatives was performed based on obtained values shown in Table 6. The results of the ranking of the final global weights of alternatives are presented in Table 7. According to the result findings, it can be seen that Contractor 4 ranked first, followed by Contractor 3 and so on.

**Table 7.** Sorting and Ranking of Final Global Weights of Alternatives

| Alternatives Names | Final Global Weights Of Alternatives | Ranking |
|---|---|---|
| Contractor 4 | 0.001374 | 1 |
| Contractor 3 | 0.001363 | 2 |
| Contractor 1 | 0.001326 | 3 |
| Contractor 8 | 0.001180 | 4 |
| Contractor 6 | 0.001108 | 5 |
| Contractor 9 | 0.000847 | 6 |
| Contractor 5 | 0.000765 | 7 |
| Contractor 2 | 0.000545 | 8 |
| Contractor 7 | 0.000332 | 9 |

### 3.4 The Implementation of the Financial Evaluation Stage

A tender guarantee which is also bid security is not compulsory on this type of project because the cost is below #300million. Thereafter the individual bid price is compared with the Consultant's figure (#143,034,460.84) and the differences are computed as can be seen in Table 8.

**Table 8.** Difference between Consultant's Figure and Alternatives Bid Prices

| Contractors | Bid Prices (Tender Sum) | Difference |
|---|---|---|
| Contractor 4 | 141,565,965.72 | -1,468,495.12 |
| Contractor 3 | 143,431,759.87 | 397,299.03 |
| Contractor 1 | 141,853,042.08 | -1,181,418.76 |
| Contractor 8 | 136,494,671.46 | -6,539,789.38 |
| Contractor 6 | 184,624,400.10 | 41,589,939.26 |
| Contractor 9 | 160,311,181.21 | 17,276,720.37 |
| Contractor 5 | 121,187,832.10 | -21,846,628.74 |
| Consultant's figure was #143,034,460.84 | | |

The seven (7) contractors that are qualified for financial evaluation can execute this project and their costs are within reasonable limits. From Table 8, it can be concluded that the lowest and most responsive contractor is Contractor 5 that returned a contract sum of #121,187,832.10 and is therefore recommended for the award of the contract.

## 4      Conclusions

The research work has established feedback integrated web-based multi-criteria group decision-making models that will help in the selection of contractors using the Fuzzy Analytic Hierarchy Process and the proposed model was implemented in a web-enabled environment. The proposed model is applied to solve the contractor selection problem of the construction of No 1 Three-in-One Lecture Theatre for Centre for Entrepreneurship at The Federal University of Technology, Akure. The results demonstrate the viability and practicability of the Fuzzy Analytic Hierarchy Process methodology to the contractor selection problem. The proposed model for contractor selection in Nigeria is expected to improve the manual tendering processes and yet make it convenient for decision-makers to manage the selection process. It will also increase the integrity and transparency of the prequalification and evaluation of tendering processes, thereby aiding timely decision making and reduce bias on the part of the decision-makers while allowing participation in an online platform. For future work, it is recommended that additional criteria and sub-criteria should be used to evaluate contractors with more case studies, more rigorous evaluation and more extensive experiments.

## References

1. Aronson, J., Myers, R., Wharton, R.: Time Pressure Impacts on Electronic Brainstorming in a Group Support System Environment. Informatica 24(2), 149-158 (2000).
2. Morais, D. C., Almeida, A. T.: Water Supply System Decision Making using Multi-Criteria Analysis. Water S.A. 32(2), 229-235 (2006).
3. Ozer, I., Lane, D.: Multi-Criteria Group Decision Making Methods and Integrated Web-Based Support Systems. In Group Decisions and Negotiations Conference, pp. 1-24. Mont Tremblant, Quebec (2010).
4. Zhang, S., Goddard, S.: A Software Architecture and Framework for Web-Based Distributed Decision Support Systems. Decision Support Systems 43(4, 1133-1150 (2007).
5. Saaty, T.: The Analytical Hierarchy Process. McGraw-Hill, New York, USA (1980)
6. Loken, E.: Use of Multi-Criteria Decision Analysis Methods for Energy Planning Problems. Renewable and Sustainable Energy Reviews 11(7), 1584-1595 (2007).
7. Chan, F., Chan, H., Ip, R., Lau, H.: A Decision Support System for Supplier Selection in the Airline Industry. Journal of Engineering Manufacture 221( 4), 741-758 (2007).
8. Zadeh, L.A.: Fuzzy Sets. Information and Control 8(3), 338-353 (1965).
9. Takacs, M.: Multilevel Fuzzy Approach to Risk and Disaster Management. Acta Polytechnica Hungarica 7(4), 91-102 (2010).
10. Zheng, G., Zhu, N., Tian, Z., Chen, Y., Sun, B.: Application of a Trapezoidal Fuzzy AHP Method for Work Safety Evaluation and Early Warning Rating of Hot and Humid Environments. Safety Science 50(2), 228-239 (2012).
11. Wang, T.: The Interactive Trade Decision-Making Research: An Application of Novel Hybrid MCDM Model. Economic Modelling 29(3), 926-935 (2012).
12. Vahidnia, M. H., Alesheikh, A., Alimohammadi, A., Bassiri, A.: Fuzzy Analytic Hierarchy Process in GIS Application. The International Archives of the Photogrammetry, Remote Sensing and Spatial Information Sciences 37, 593-596 (2008).
13. Iranmanesh, H., Shirkouhi, S., Skandari, M.: Risk Evaluation of Information Technology Projects based on Fuzzy Analytic Hierarchy Process. International Journal of Computer and Information Science and Engineering 2(1), 38-44 (2008).
14. Srdjevic, B., Medeiros, Y.: Fuzzy AHP Assessment of Water Management Plans. Water Resources Management 22(7), 877-894 (2008).
15. Li, L., Shi, Z., Yin, W., Zhu, D., Ng, S., Cai, C., Lei, A.: A Fuzzy Analytic Hierarchy Process Approach to Eco-environmental Vulnerability Assessment for the Danjiangkou Reservoir Area, China. Ecological Modelling 220(23), 3439-3447 (2009).
16. Adebiyi, A., Ayo, C., Adebiyi, M.: Development of Electronic Government Procurement System for Nigeria Public Sector. International Journal of Electrical and Computer Sciences IJECS-IJENS, 71-84 (2010).

17. Simon, L., Jayakumar, A.: A Multi-Perspective Strategic Approach for the Selection of Contractor in Highway Projects. International Journal of Innovative Research in Science, Engineering and Technology 5(2), 2402-2407 (2016).
18. Rahmatollah, G., Gholamreza, J., Reza, R..: Contractor Selection in MCDM Context using Fuzzy AHP. Iranian Journal of Management Studies (IJMS) 7(1), 151-173 (2014).
19. Khan, M., Arvind, J., Veepan, k.: Multi-Criteria Supplier Selection Using Fuzzy-AHP Approach: A Case Study of Manufacturing Company. International Journal of Research in Mechanical Engineering & Technology 5(1), 73-79 (2015).
20. Tran, T. H., Long, L. H., Young, D. L.: A Fuzzy AHP Model for Selection of Consultant Contractor in Bidding Phase in Vietnam. KICEM Journal of Construction Engineering and Project Management 5(2), 35-43 (2015).
21. Burney, A.S.M., Ali, S.M.: Fuzzy Multi-Criteria Based Decision Support System for Supplier Selection in Textile Industry. International Journal of Computer Science and Network Security 19(1), 239-244 (2019).
22. Kaufmann, A., Gupta, M.: Fuzzy Mathematical Models in Engineering and Management Science, Elsevier Science Inc, Netherlands (1988).
23. Lamata, M.: Domestic Wastewater Treatment in Developing Countries. Earthscan Publications, U.K. (2004).
24. Buckley, J. J.: Fuzzy Hierarchical Analysis. Fuzzy sets and systems 17(3), 233-247 (1985).
25. Ayhan, M. B.: A Fuzzy AHP Approach for Supplier Selection Problem: A Case Study in a Gearmotor Company. International Journal of Managing Value and Supply Chains, 4(3), 11-23 (2013).
26. Chou, S.W., Chang, Y.C.: The Implementation Factors that influence the Enterprise Resource Planning Benefits. Decision Support Systems 46(1), 149-157 (2008).
27. Saaty, T.: Decision Making for Leaders: The Analytic Hierarchy Process for Decisions in a Complex World, 5th edn. Pittsburg (2012).
28. Cabala, P.: Using the Analytic Hierarchy Process in Evaluating Decision Alternatives. Operations Research and Decisions 1, 1-23 (2010).
29. Herrera-Viedma, E., Martínez, L., Mata, F., Chiclana, F.: A Consensus Support System Model for Group Decision-Making Problems with Multi-Granular Linguistic Preference Relations. IEEE Transactions on Fuzzy Systems 13, 644–645 (2005).